# Paper title: Predicting Barge Presence and Quantity on Inland Waterways using Vessel Tracking Data: A Machine Learning Approach

Geoffery Agorku[a*], Sarah Hernandez[a], Maria Falquez[a], Subhadipto Poddar[a], Shihao Pang[a]

[a] Department of Civil Engineering, University of Arkansas, Fayetteville, Arkansas, 7270, USA

* Corresponding author.

Email Address: gagorku@uark.edu

**Acknowledgments**

The authors acknowledge the support and sponsorship provided by the National Science Foundation (NSF), Directorate of Engineering, Division of Civil, Mechanical, and Manufacturing Innovation [Award number 2042870], and the U.S. Army Corps of Engineers (USACE), Engineer Research and Development Center (ERDC), Coastal Hydraulics Laboratory (CHL).

**Abstract**

This study presents a machine learning approach to predict the number of barges transported by vessels on inland waterways using tracking data from the Automatic Identification System (AIS). While AIS tracks the location of tug and tow vessels, it does not monitor the presence or number of barges transported by those vessels. Understanding the number and types of barges conveyed along river segments, between ports, and at ports is crucial for estimating the quantities of freight transported on the nation's waterways. This insight is also valuable for waterway management and infrastructure operations impacting areas such as targeted dredging operations, and data-driven resource allocation. Labeled sample data was generated using observations from traffic cameras located along key river segments and matched to AIS data records. A sample of 164 vessels representing up to 42 barge convoys per vessel was used for model development. The methodology involved first predicting barge presence and then predicting barge quantity. Features derived from the AIS data included speed measures, vessel characteristics, turning measures, and interaction terms. For predicting barge presence, the AdaBoost model achieved an F1 score of 0.932. For predicting barge quantity, the Random Forest combined with an AdaBoost ensemble model achieved an F1 score of 0.886. Bayesian optimization was used for hyperparameter tuning. By advancing predictive modeling for inland waterways, this study offers valuable insights for transportation planners and organizations, which require detailed knowledge of traffic volumes, including the flow of commodities, their destinations, and the tonnage moving in and out of ports.

## 1. Introduction

The movement of freight on inland waterways is a critical component of the global supply chain, providing an efficient and environmentally friendly alternative to road and rail transport (Wang et al., 2024). In the United States, the inland waterway system spans over 12,000 miles, transporting significant quantities of bulk commodities such as agricultural products, coal, and petroleum (Satar & Peoples, 2010). Optimizing this transport mode requires accurate and timely data on barge movements, which has traditionally been challenging to obtain (Sugrue & Adriaens, 2021).

The integration of various data sources has become increasingly important in monitoring maritime transportation including the Automatic Identification System (AIS) and Lock Performance Monitoring System (LPMS) (Guo et al., 2023; Sugrue & Adriaens, 2021). AIS technology provides real-time data on vessel movements (H. Li & Yang, 2023). While AIS provides the position of vessels like tow and tug boats, it does not track the presence or quantity of barges towed by vessels, a crucial parameter for understanding waterway traffic dynamics and informing decisions on waterway management and infrastructure investment (Sugrue & Adriaens, 2021). Sources such as the LPMS (*Locks*, n.d.) provide commodity and barge counts but correspond only to the location of locks, not all river segments. Cameras coupled with computer vision can also be used to monitor barge volumes but are limited in spatial coverage (Agorku et al., 2024).

This study proposes a novel approach to predict the quantity of barges transported by vessels on inland waterways by applying machine learning to AIS-derived vessel maneuverability features. Labeled sample data for training, validation, and testing is generated from publicly available traffic cameras with opportune view angles of inland waterway segments and matched to respective AIS records. Features for the machine learning models include rate of turn, speed over ground, and vessel characteristics, among others. The model is applicable to real-time prediction of barge presence and quantity. Reliable barge count data can inform the scheduling of lock operations, reduce congestion, and improve waterway transport efficiency (Bu & Nachtmann, 2023; Shobayo & van Hassel, 2019). The objectives of this paper are as follows:

1. Develop robust predictive models using machine learning techniques to accurately determine whether vessels transport barges, and

2. For vessels transporting barges, predict the quantity of barges being transported.

## 2. Background

Previous studies explore the application of AIS data for coordinating river port operations, vessel trajectory prediction, collision risk assessment, anomaly detection, predicting shipping efficiency, and supporting real-time carbon accounting for maritime transport (Z. Li et al., 2024; Sugrue & Adriaens, 2021; Wang et al., 2024; Y. Yang et al., 2024). Despite these applications, a significant gap remains in predicting the quantity and type of cargo carried by vessels on waterways (Sugrue & Adriaens, 2021). Addressing this gap presents an opportunity for enhancing waterway management through advanced predictive analysis.

The LPMS (*Locks*, n.d.) and Waterborne Commerce Statistics Center (WCSC) (*WCSC Waterborne Commerce Statistics Center*, n.d.) are notable systems for monitoring barge volumes. LPMS collects and reports data on vessels traversing US Army Corps of Engineers (USACE)-maintained locks (*Corps Locks - Home*, n.d.). While LPMS data is valuable, it does not account for vessel movements that do not pass through USACE - owned locks, and experiences a lag of one month, limiting its timeliness and comprehensiveness. Similarly, WCSC collects, processes, distributes, and archives vessel trip and cargo data for ports (*WCSC Waterborne Commerce Statistics Center*, n.d.). These statistics are used to produce monthly indicators and an annual report on waterborne commerce (*WCSC Waterborne Commerce*, n.d.). While WCSC data offer extensive origin-to-destination information, they often face a significant delay—typically around 2–3 years—due to lengthy acquisition and verification processes, and instances of non-reporting. This delay limits the utility of WCSC data for real-time decision-making, planning, and developing metrics for waterway management. (*Waterborne Commerce Monthly Indicators Available to Public*, n.d.) . Furthermore, reliance on vessel operators' submission practices to LPMS and WCSC can result in delays, omissions, or inaccuracies (Asborno et al., 2020). Our study aims to address these gaps by exploring the feasibility of real-time data collection mechanisms, which can offer immediate insights while also complementing existing datasets.

The expanding network of AIS receivers, coupled with advancements in analytical methods, has enhanced the utility of AIS data across various applications (Pallotta et al., 2013). This growth not only broadens coverage but also enables sophisticated tracking of vessels, providing comprehensive insights into maritime activities and supporting navigation and safety measures (Mallikarjuna Reddy & Anuradha, 2020). AIS data, which includes information on vessel identity, position, speed, and cargo, is used for maritime traffic management (Asborno et al., 2022; Z. Li et al., 2024; D. Yang et al., 2021). However, AIS does not capture the presence or quantity of barges being towed by vessels, a critical parameter for quantifying waterway traffic dynamics and making informed decisions regarding waterway management and infrastructure investment (Sugrue & Adriaens, 2021).

Illustrating the potential of AIS data application regarding vessel tracking, Asborno et al. proposed a map-matching algorithm for inland waterways (Asborno et al., 2022). The study evaluated AIS data observed at the McClellan-Kerr Arkansas River Navigation System (MKARNS) and the Mississippi River along Arkansas's border using 7,803,151 AIS records broadcasted by 776 vessels. The algorithm identified 120,185 stops with an 84.0% accuracy and 47,555 trips with an 83.5% accuracy, demonstrating precise tracking of vessel movements. Similarly, Pallotta et al. proposed the Traffic Route Extraction and Anomaly Detection (TREAD) methodology, employing a density-based clustering algorithm for extracting vessel routes and detecting anomalies using AIS data. Tested in areas like the Strait of Gibraltar and the North Adriatic Sea, TREAD achieved learning accuracy of up to 95% in high-traffic areas (Pallotta et al., 2013). Capobianco et al. proposed vessel trajectory prediction models based on sequence-to-sequence approaches using encoder-decoder recurrent neural networks (RNNs), specifically incorporating Long Short-Term Memory (LSTM) RNNs. These models, trained on historical AIS data, predict future vessel trajectories, capturing complex space-time dependencies. Notably, the labeled encoder-decoder model with an attention aggregation function achieved a final displacement error below 2.5 nautical miles in approximately 90% of the cases (Capobianco et al., 2021). These studies show that machine learning algorithms can effectively forecast vessel movement as trajectories, stops, and trips. However, all focus on the vessel, rather than what the vessel is transporting. Building on prior AIS data research, there is an opportunity to extend methodologies to predict the quantity and types of barges carried by vessels on waterways.

## 3. Methods

The methodology is presented in two parts: (1) data preprocessing and (2) model selection and development.

### *3.1. Data Preprocessing*

Data preprocessing included cleaning the AIS data, matching AIS records to a representative inland waterway network, and reconstructing vessel trajectories.

### *3.2. Cleaning AIS Data*

Records with zero speed were removed. Some of these records resulted from vessel operators unintentionally failing to turn off AIS equipment after completing trips (Wolsing et al., 2022). A spatial buffer was created around each river channel to filter out AIS data points falling outside the navigable waterways. This step ensured that only relevant data points within the river boundaries were retained.

### *3.3. Map Matching*

Vessel trajectories may be incomplete such that the same vessel is seen in disconnected segments of the river. For example, missing records approximately 40 miles from the Emerson River Bridge (ERB) lead to a disconnected trajectory **(Figure 1)**. A map-matching process is required to reconstruct such trajectories.

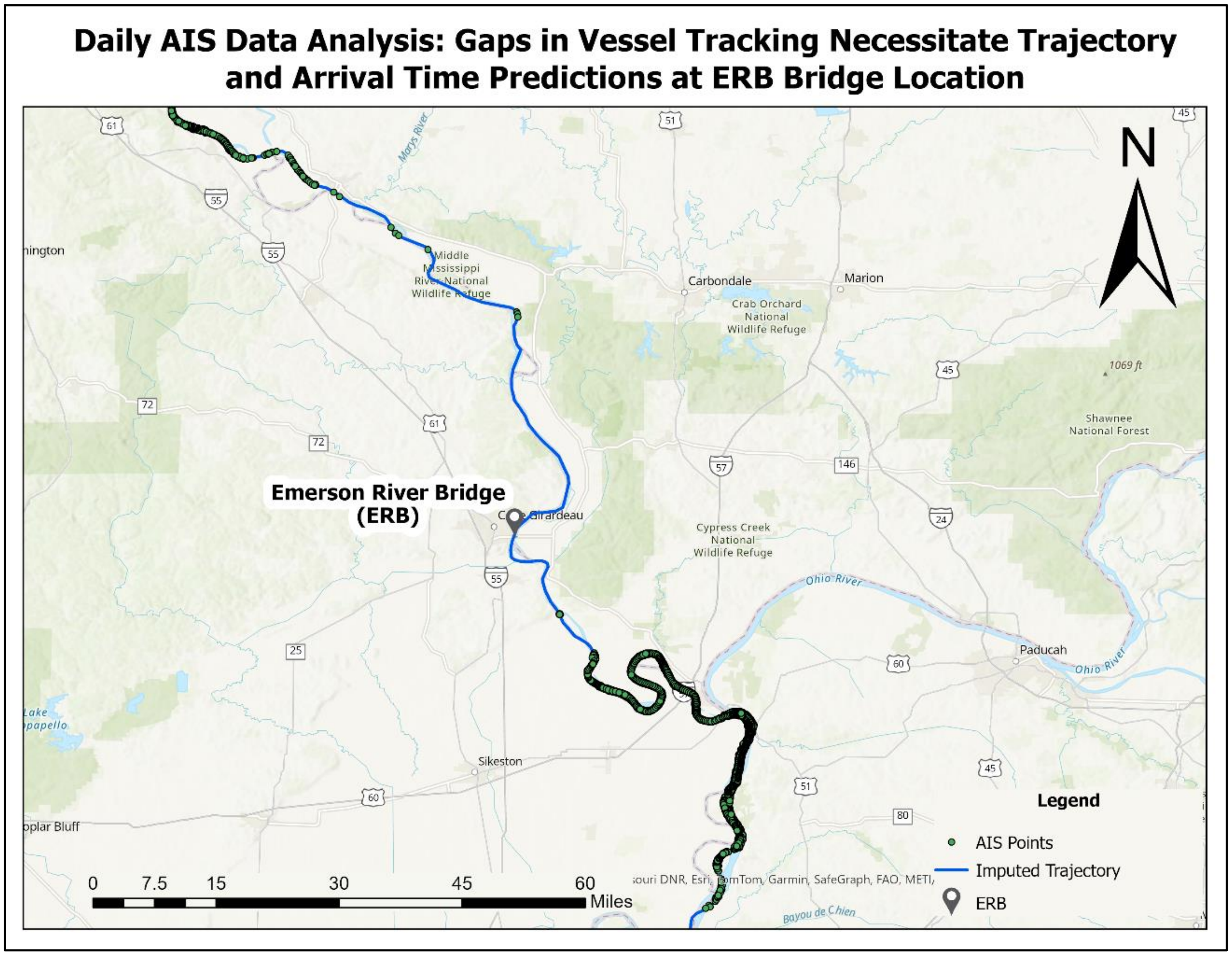

Figure 1: Example of daily analysis for ERB location showing missing records for 2023-04-08

The method for predicting missing vessel trajectories builds upon Asborno et al. who applied map-matching heuristics to AIS data, identifying vessel stops by clustering consecutive position records based on location, timestamp, and speed. Each stop was associated with the nearest network node (e.g., links, locks, port terminals, and barge staging areas) (Asborno et al., 2022). The algorithm calculated the shortest path between nodes for consecutive stops, reconstructing the vessel's trajectory by connecting these path segments. Asborno et al.'s approach lacked the necessary granularity for our analysis. It associated AIS vessel pings with nodes potentially far apart and found the shortest path between them, assuming vessels always take the shortest route between consecutive stops. A more granular approach was introduced to address these shortcomings by dividing the river into smaller segments and calculating the center of gravity (COG) of points for each segment. The COG in this context refers to the average position of all points within the river segment, representing the balanced midpoint for that segment. The river was divided into equal segments. The segment length necessary to balance computational complexity and accuracy was learned through sensitivity analysis described in section 4. For each segment, the COG of AIS data points belonging to all vessels was calculated using the following steps:

a. The distance from each AIS data point within a segment to all other points within the same segment was calculated.
b. The average of all points within a segment was computed to determine the COG. This ensured that the COG accurately represented the average position of vessels within the segment.

This process was repeated for all segments along the river. This average path was then used to impute AIS trajectories when needed, allowing for accurate pairing of vessel trajectories from AIS data with vessel images from traffic cameras to generate a labeled sample dataset for model development and evaluation.

### *3.4. Model Development*

Model development includes model selection, feature engineering, feature selection, and data augmentation.

### *3.4.1. Model Selection*

In total, 10 models were evaluated representing base learners (3 models), ensembles (4), and hybrid ensembles (3). Base models include Support Vector Machines (SVM) (*Support Vector Machines | IEEE Journals & Magazine | IEEE Xplore*, n.d.), K-Nearest Neighbor (KNN) (Peterson, 2009), and Neural Networks (NN) (*Neural Networks and Their Applications | Review of Scientific Instruments | AIP Publishing*, n.d.). Basic ensemble methods include: Random Forests (RF) (Breiman, 2001), XGBoost (Extreme Gradient Boosting) (*XGBoost | Proceedings of the 22nd ACM SIGKDD International Conference on Knowledge Discovery and Data Mining*, n.d.), AdaBoost (Adaptive Boosting) (Schapire, 2013), and LightGBM (Light Gradient-Boosting Machine). Hybrid ensembles include RF+AdaBoost, RF+AdaBoost+LightGBM, and RF + LightGBM. Models were chosen for their effectiveness in similar tasks and ability to handle heterogeneous datasets.

NNs model complex, non-linear relationships through their deep, layered architecture (Malfa et al., 2024). SVMs seek optimal hyperplanes in high-dimensional spaces (Blanco et al., 2020). KNNs capture local patterns based on proximity (Anava & Levy, 2016). Decision Trees handle both numerical and categorical data through a hierarchical structure of splits (Gunluk et al., 2019). These models were applied on their own and used as base learners in the following ensemble approaches. Ensemble approaches combine the results of multiple base learners through mechanisms like voting, weighting, etc. Basic ensemble models like RF reduce overfitting through the aggregation of multiple decision trees, enhancing accuracy and providing feature importance measures (Iranzad & Liu, 2024). AdaBoost iteratively adjusts the weights of misclassified instances to improve the performance of weak learners (Beja-Battais, 2023). LightGBM uses histogram-based algorithms and leaf-wise tree growth, allowing it to handle large datasets with higher efficiency and faster training times (Ke et al., 2017). XGBoost uses regularization techniques that prevent overfitting. [34]. All the basic ensemble models used decision trees as base models.

Each model in the hybrid ensemble independently makes predictions on the same training set, producing a new training set composed of these predictions (*StackGenVis: Alignment of Data, Algorithms, and Models for Stacking Ensemble Learning Using Performance Metrics | IEEE Journals & Magazine | IEEE Xplore*, n.d.).

This new training set is then used to train a meta-model, which aims to capture patterns and relationships between the base model predictions. The meta-model ultimately produces the final predictions (Huang et al., 2022) **(Figure 2).** Model stacking leverages the strengths and compensates for the weaknesses of individual models, enhancing overall predictive performance.

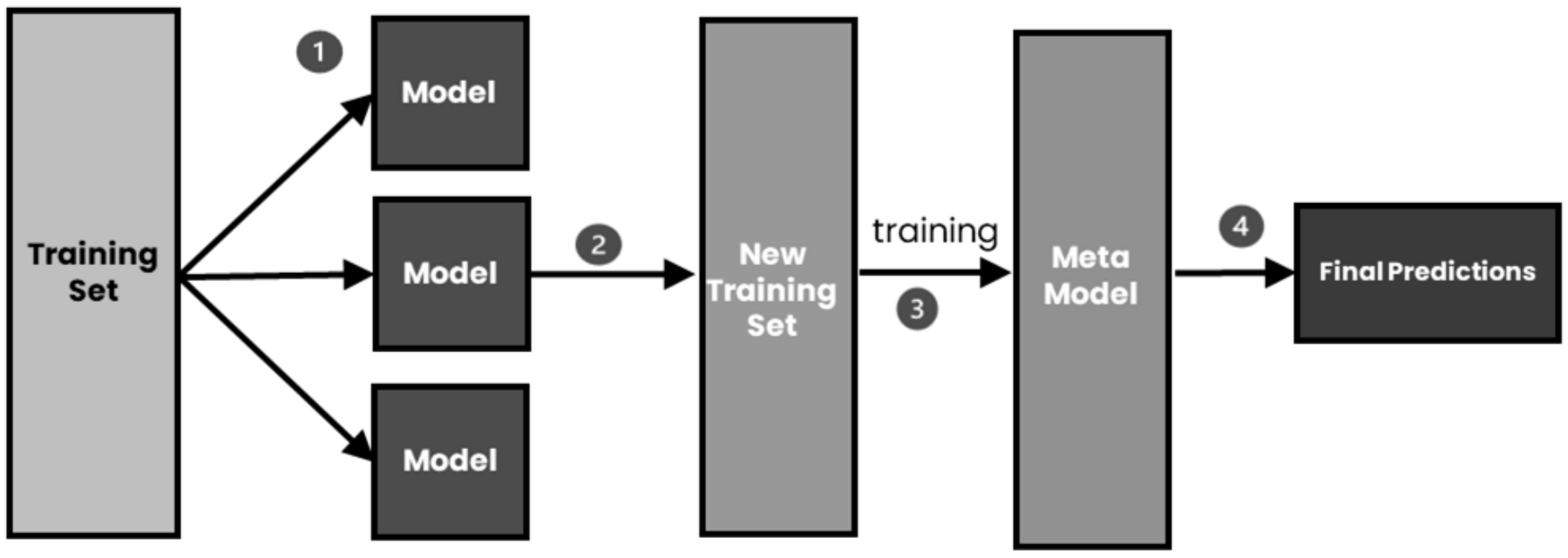

Figure 2: Illustration of the Model Stacking Process (Soni, 2023)

### *3.4.2. Feature Engineering*

36 categorical and numerical features were derived from AIS data to capture vessel operational characteristics assumed to influence barge transport. The features include speed, maneuverability, vessel type, cargo type, dimensions, draft, acceleration, and operational status **(Table 1).**

Table 1: AIS-Derived Features for Barge Prediction

| # | Feature Group | Description | Feature(s) | Acronym |
|---|---|---|---|---|
| 1 | Speed Over Ground (SOG) | Vessel movement speed; Barge count affects vessel speed (fewer barges = higher SOG; more barges = lower SOG) | Lower Quartile SOG | SOG_Q1 |
| | | | Median SOG | SOG_Q2 |
| | | | Upper Quartile SOG | SOG_Q3 |
| | | | Percentage of Time Spent Traveling | PTST_SOG _<5.4 |

| # | Feature Group | Description | Feature(s) | Acronym |
|---|---|---|---|---|
| | | | (PTST) less than 5.4 knots | |
| | | | PTST Between SOG 5.4 and 6.9 knots | PTST_SOG_ 5.4_to_6.9 |
| | | | PTST at SOG more than 6.9 knots | PTST_SOG _>6.9 |
| | | | Standard Deviation of SOG | SOG_SD |
| | | | Square of Median Speed | (SOG_Q2)^2 |
| | | | Square of Standard Deviation of SOG | (SOG_SD)^2 |
| | | | Median SOG Squared | (SOG_Q2)^2 |
| | | | Median SOG cubed | (SOG_Q2)^3 |
| 2 | Normalized Rate of Turn (NROT) | Average heading change per minute, adjusted for data frequency; Higher ROT requires more maneuverability, limiting barge control in narrow or curving channels | Normalized Rate of Turn | NROT |
| 3 | Vessel Type | Vessel classification in AIS data; Vessel type affects towing capacity and operational constraints, influencing the number of barges a | Vessel Type 31 (Towing) | VT_31_Towing |
| | | | Vessel Type 52 (Tug) | VT_52_Tug |
| | | | Vessel Type Other (Not Tug or Tow) | VT_Other |

| # | Feature Group | Description | Feature(s) | Acronym |
|---|---|---|---|---|
| | | vessel can tow based on size, power, and design | | |
| 4 | Cargo Type | Classification of transported goods; Cargo type affects barge quantity due to varying weight, volume, and stability requirements | Cargo Type 31 (Towing) | CT_31 |
| | | | Cargo Type 32 (Towing: length exceeds 200m or breadth exceeds 25m) | CT_32 |
| | | | Cargo Type 52 (Tug) | CT_52 |
| | | | Cargo Type 57 (Spare - for assignment to local vessel) | CT_57 |
| | | | Cargo Type Other (Not 31,32,52, or 57) | CT_Other |
| 5 | Vessel Dimensions | Length and width of a vessel; Larger vessels can manage more barges while ensuring stability and navigability (Duldner-Borca et al., 2024) | Length | Len |
| | | | Width | Wid |
| | | | Length*Width | Len*Wid |
| | | | Length* Median SOG | Len* SOG_Q2 |
| | | | Width* Median SOG | Wid* SOG_Q2 |
| | | | Length*Width* Median SOG | Len*Wid*SOG_Q2 |
| | | | Length Squared | (Len)^2 |
| | | | Width Squared | (Wid)^2 |
| | | | Length Cubed | (Len)^3 |

| # | **Feature Group** | **Description** | **Feature(s)** | **Acronym** |
|---|---|---|---|---|
| 6 | Draft | Depth of a vessel below the waterline; Deeper draft limits navigation in shallow waters, reducing barge towing capacity. Draft in AIS data is static. | Vessel Draft | VDraft |
| 8 | Acceleration | Rate of speed change; Acceleration reflects vessel behavior under various loads | Standard Deviation of Acceleration | Acc_SD |
| | | | Standard Deviation of Acceleration * Standard Deviation of Speed | Acc_SD*SOG_SD |
| 9 | Operational Status | Current activity or condition of a vessel; Operational status may affect barge capacity due to varying power, towing methods, etc | Status 0 (Underway using engine) | Status_0 |
| | | | Status 12 (power-driven vessel pushing ahead or towing alongside) | Status_12 |
| | | | Status 15 (undefined = default) | Status_15 |
| | | | Status Other (Not 0,12 or15) | Status_Other |

*3.4.3. Feature Selection*

Feature selection algorithms such as Recursive Feature Elimination (RFE) identify the most relevant features while eliminating those that are less significant (Jahed & Tavana, 2024). Initially, a machine learning model is trained on all features, and feature importance scores are evaluated. The least important features are iteratively removed, and the model is retrained on the reduced feature set. This process continues until the optimal number of features is reached, balancing model complexity and performance (*Applied Sciences | Free Full-Text | Hybrid-Recursive Feature Elimination for Efficient Feature Selection*, n.d.).

### *3.5. Model Training and Validation*

A hierarchical methodology was used to first predict the presence of barges and then quantify the number of barges in tow (**Figure 3**). A binary classification was used for the barge presence model: [Vessel with Barge; Vessel Without Barge]. A six-class scheme was used for the barge quantity model: [1 Barge, 2-4 Barges, 5-12 Barges, 13-20 Barges, 21-29 Barges, >29 Barges (30-42 Barges)]. A sensitivity analysis was performed to determine barge class groups and is described in section 4.7.2.

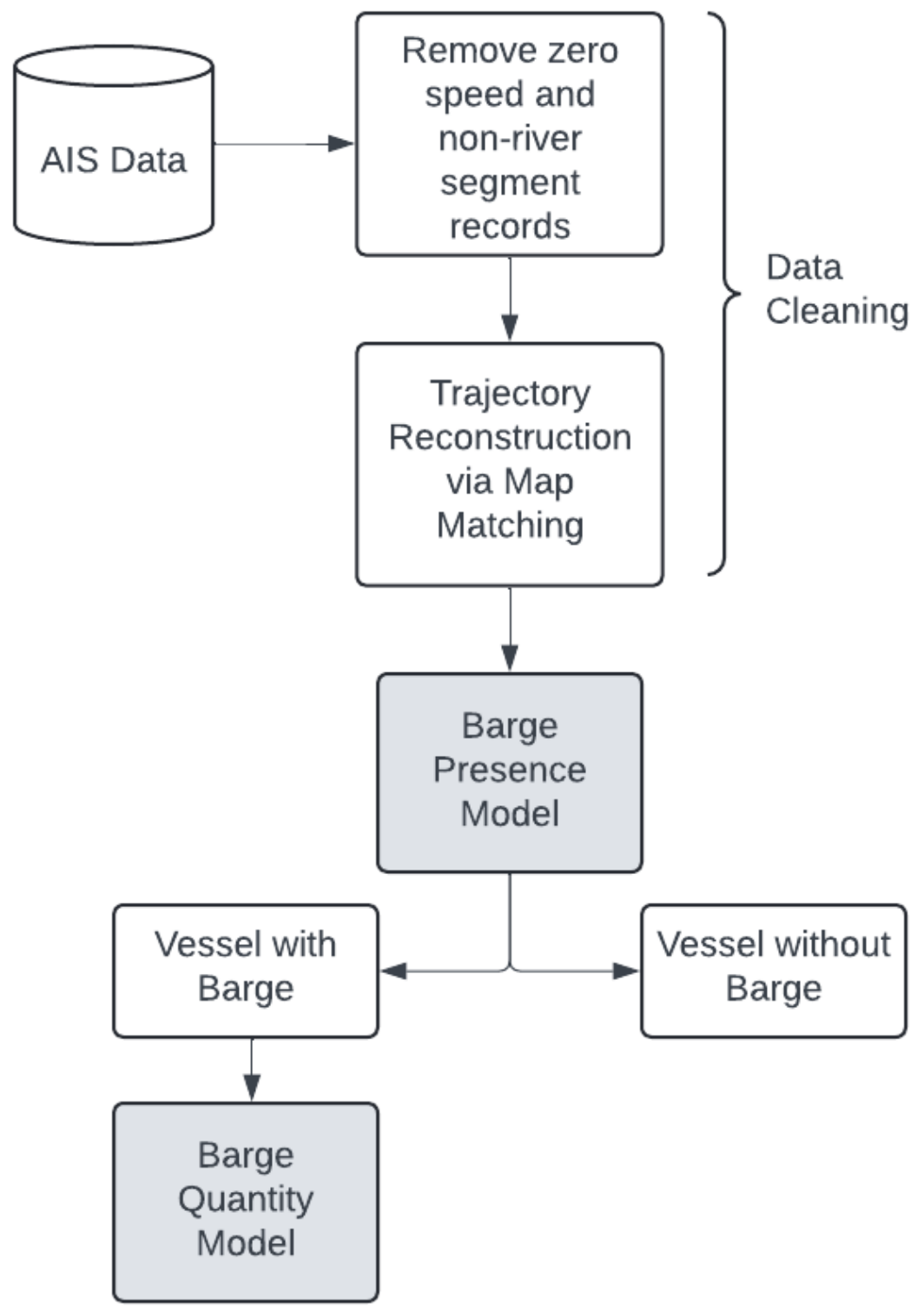

Figure 3: Hierarchical Model for Barge Presence and Barge Quantity Estimation including Data Cleaning

### *3.6. Model Evaluation*

Accuracy, precision, recall, F1 score, and ROC-AUC were used to assess model performance. In addition, a confusion matrix is presented to visualize cross-classification errors.

1. **F1 Score:** The harmonic means of precision and recall ideal for classification problems with imbalanced datasets **(Equation 3).**

$$Precision = \frac{True\ Positives\ (TP)}{True\ Positives\ (TP) + False\ Positives\ (FP)} \quad \textbf{Equation 1}$$

$$Recall = \frac{True\ Positives\ (TP)}{True\ Positives\ (TP) + False\ Negatives\ (FN)} \quad \textbf{Equation 2}$$

$$F1\ Score = 2 \times \frac{Precision\ \times Recall}{Precision + Recall} \quad \textbf{Equation 3}$$

2. **Accuracy:** The ratio of correctly predicted instances to total instances **(Equation 4).**

$$Accuracy = \frac{True\ Positives\ (TP) + True\ Negatives\ (TN)}{Total\ Number\ of\ Instances}$$ **Equation 4**

3. **ROC-AUC Score**: Measures the model's ability to distinguish between classes, with higher values indicating better performanc**e (Equation 5).**

$$ROC - AUC = \int_0^1 TPR(FPR)d(FPR)$$ **Equation 5**

TPR = True Positive Rate (Recall); FPR = False Positive Rate

### *3.5. Hyperparameter Optimization*

Hyperparameter tuning for the selected models was conducted using Optuna's Bayesian optimization (Gu et al., 2021). Bayesian optimization models the performance of the hyperparameters and uses this model to evaluate the next set of hyperparameters, leading to faster convergence and better results. The primary objective was to minimize error. The process began with an initial set of hyperparameters, either randomly selected or based on prior knowledge, to train the model and evaluate performance via the F1 score. Hyperparameter optimization was performed using stratified five-fold cross-validation on the training data to ensure robustness and generalization across data subsets. Stratifying the folds maintained class distribution, crucial for imbalanced datasets.

## 4. Results and discussion

Data from four bridge locations along the Mississippi River was used to generate labeled samples for model training, validation, and testing. The process to gather, label, and prepare (data imputation and augmentation) the labeled sample data and AIS data is described in this section (**Figure 4**).

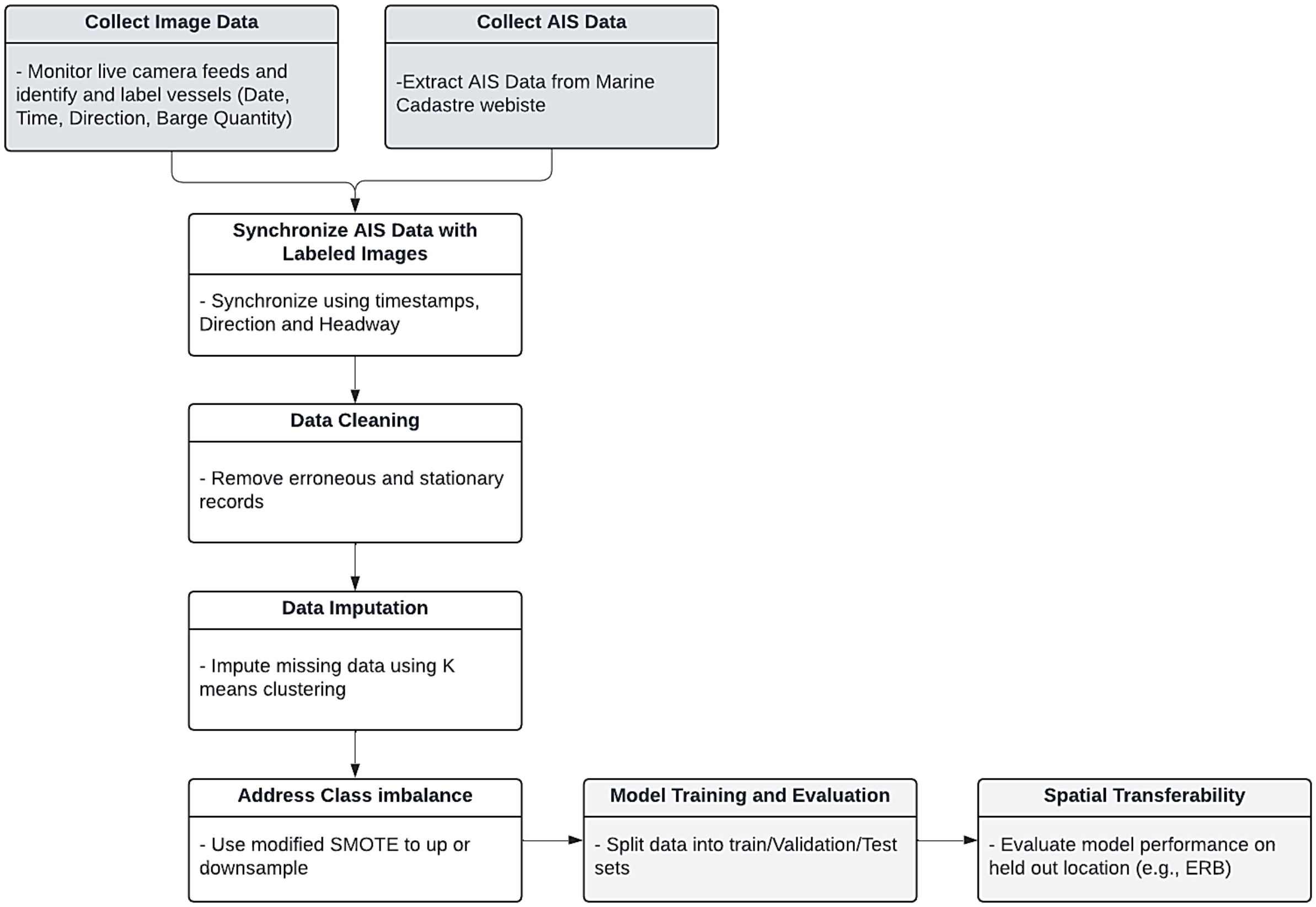

Figure 4: Sample Data Labeling Process

### *4.1. Study Location and Labeled Sample Generation*

AIS data was coupled with manual observations of vessels to create a labeled dataset for model development and evaluation. Manual observations were gathered using publicly available traffic cameras, typically maintained by state Departments of Transportation (DOTs) (*MDOTtraffic | Powered by MDOT*, n.d.; *MoDOT Traveler Information Map*, n.d.) and third-party providers like EarthCam (*EarthCam - Webcam Network*, n.d.), positioned on bridges along inland waterways. Four cameras were selected from a review of 20 cameras, considering spatial bias and resource constraints. The chosen cameras are located at the St. Louis Arch (SLA) and ERB in Missouri, the Mississippi River Bridge (MRB), and the Louisiana River Bridge (LRB) in Mississippi **(Figure 5-7).** Live camera feeds were monitored passively from October 2022 to December 2023. Manual identification of vessels and barges was done to produce labeled images that included date, time, travel direction, and quantity of barges for 419 vessel records.

The Marine Cadastre web-based tool (*AccessAIS - MarineCadastre.Gov*, n.d.) was used to extract AIS data within user-defined areas. Bounding boxes were drawn around the camera view areas to capture vessel trajectories. The study area included 811 miles of river segment: 429 miles for the upper Mississippi segment, and 382 miles for the lower Mississippi **(Figure 5).**

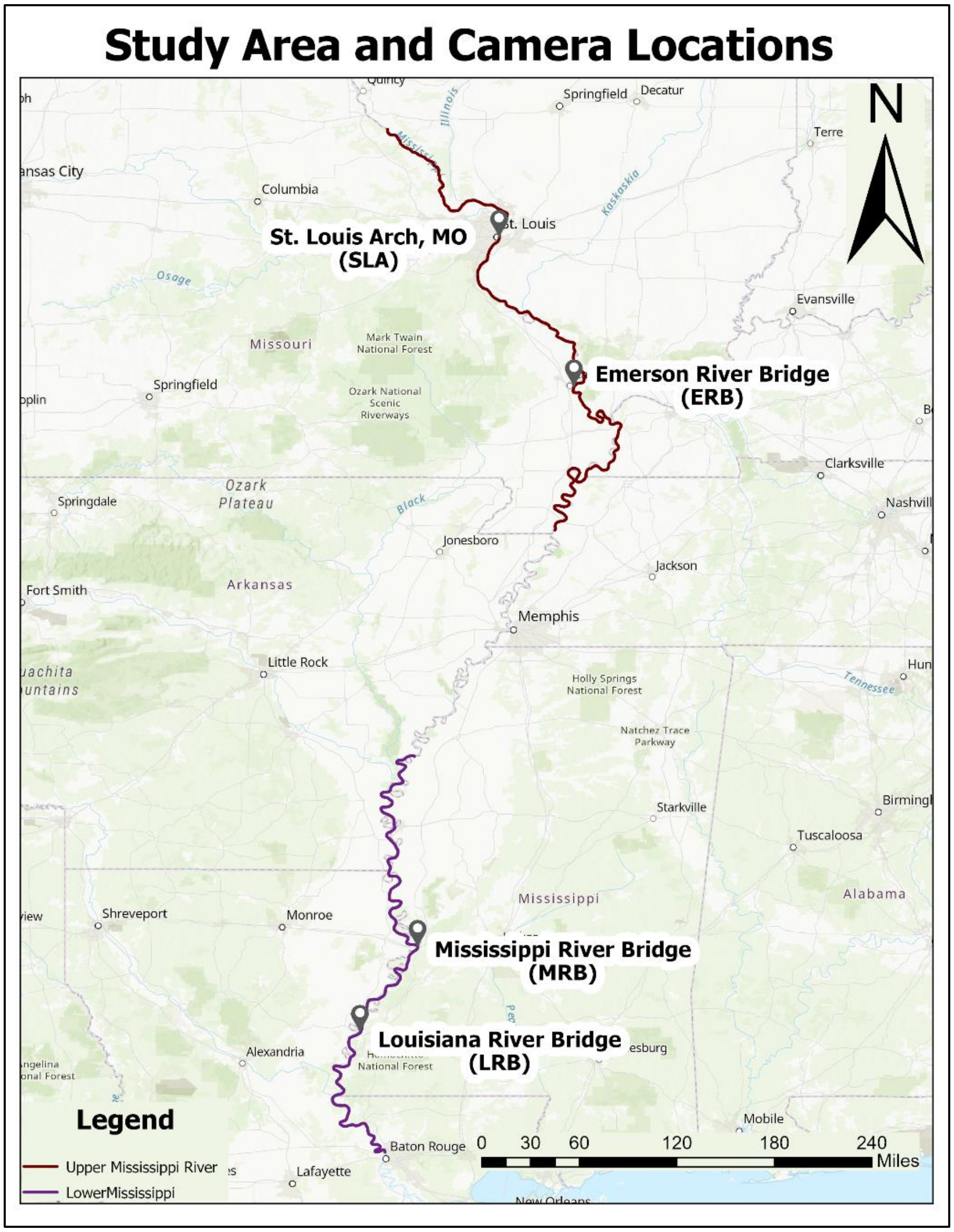

Figure 5: Locations of Traffic Cameras used for Data Collection

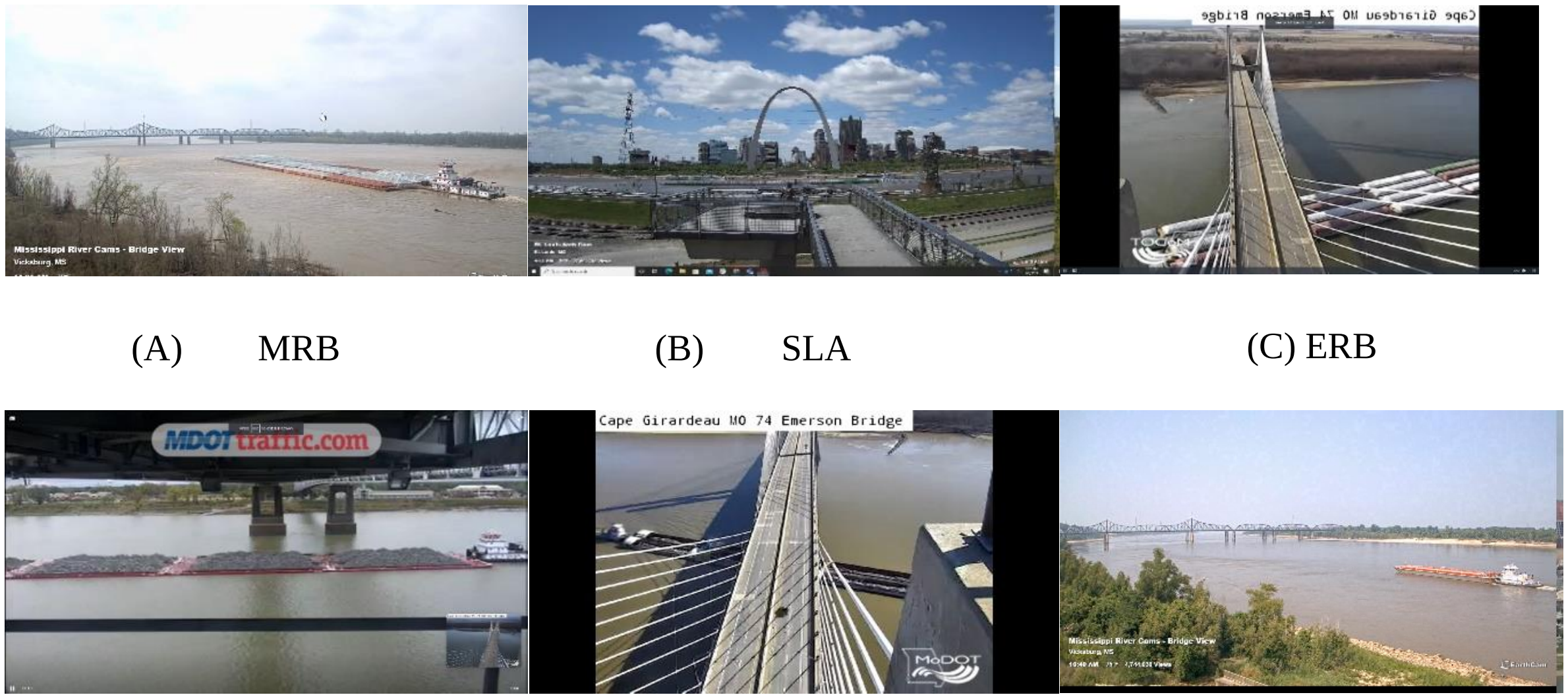

(A) MRB (B) SLA (C) ERB

(D) LRB (E) CCB (F) MRB

Figure 6: Examples of Camera Images Used for Labeled Data Generation

### *4.2. Matching AIS and Labeled Images*

While the camera feeds were available in real-time, AIS data is not available in real-time for public consumption. Therefore, AIS data and the labeled images needed to be matched to produce a labeled sample dataset for model development and evaluation. In some cases, the vessel's position recorded by AIS corresponded to the image timestamp producing a direct match of AIS record to labeled image. In many cases, however, a vessel's timestamped location from AIS was observed before and after the bridge where the image was collected, necessitating a procedure to match the vessel's AIS record to the labeled image. To accurately estimate when the vessel crossed the camera view (at the bridge), the AIS record closest to the camera/bridge location was identified. This record provided the vessel's nearest known position relative to the bridge. The distance from the vessel to the bridge was calculated using the Haversine formula **(Equation 6)** (Robusto, 1957) which is ideal for measuring distances between two points on the Earth's surface due to its consideration of the planet's spherical shape. The arrival time was computed as a weighted average of the two AIS records closest to the camera/bridge, with weights inversely proportional to their distances from the bridge. However, this method was less accurate compared to using the single closest point.

$$a = \sin\left(\frac{\Delta\phi}{2}\right) + \cos(\phi_1)\cos(\phi_2)\sin^2\left(\frac{\Delta\lambda}{2}\right) \quad \textbf{Equation 6}$$

$$c = 2\,atan\,2\left(\sqrt{a}, \sqrt{1-a}\right)$$

$$d = RC$$

where:

- $\phi_1$ and $\phi_2$ are the latitudes of the two points (in radians),
- $\lambda_1$ and $\lambda_2$ are the longitudes of the two points (in radians),
- $\Delta\phi = \phi_1 - \phi_2$ is the difference in latitude,
- $\Delta\lambda = \lambda_1 - \lambda_2$ is the difference in longitude,
- $d$ is the distance between the two points.

Finally, synchronization between the AIS records and camera images was performed using three key attributes: timestamps, direction of travel, and headway between arrivals. A match required all three attributes to align between AIS and camera data. Records with discrepancies in any attribute were not considered matches and were not used for model development. Of the 419 total camera images collected, 166 labeled images were matched to AIS vessel records (**Figure 7)**. Match percentages varied, with LRB having the highest at 81.2% and MRB the lowest at 18.6%. The low match percentages observed in certain regions, such as the MRB where significant time was invested in data collection, can be attributed to the limited availability of publicly accessible AIS records due to privacy and security constraints (Levy et al., 2023). Had the limitations of this location been known from the outset, efforts could have been adjusted to other locations to mitigate these issues and improve the match rates. Additionally, it was assumed that vessels would travel at a constant speed from their last recorded positions to the camera/bridge location. This assumption may not always hold, especially for vessels whose last recorded positions are at significant distances from the bridge. In such cases, discrepancies between the predicted arrival times of the AIS data and the timestamped images are introduced.

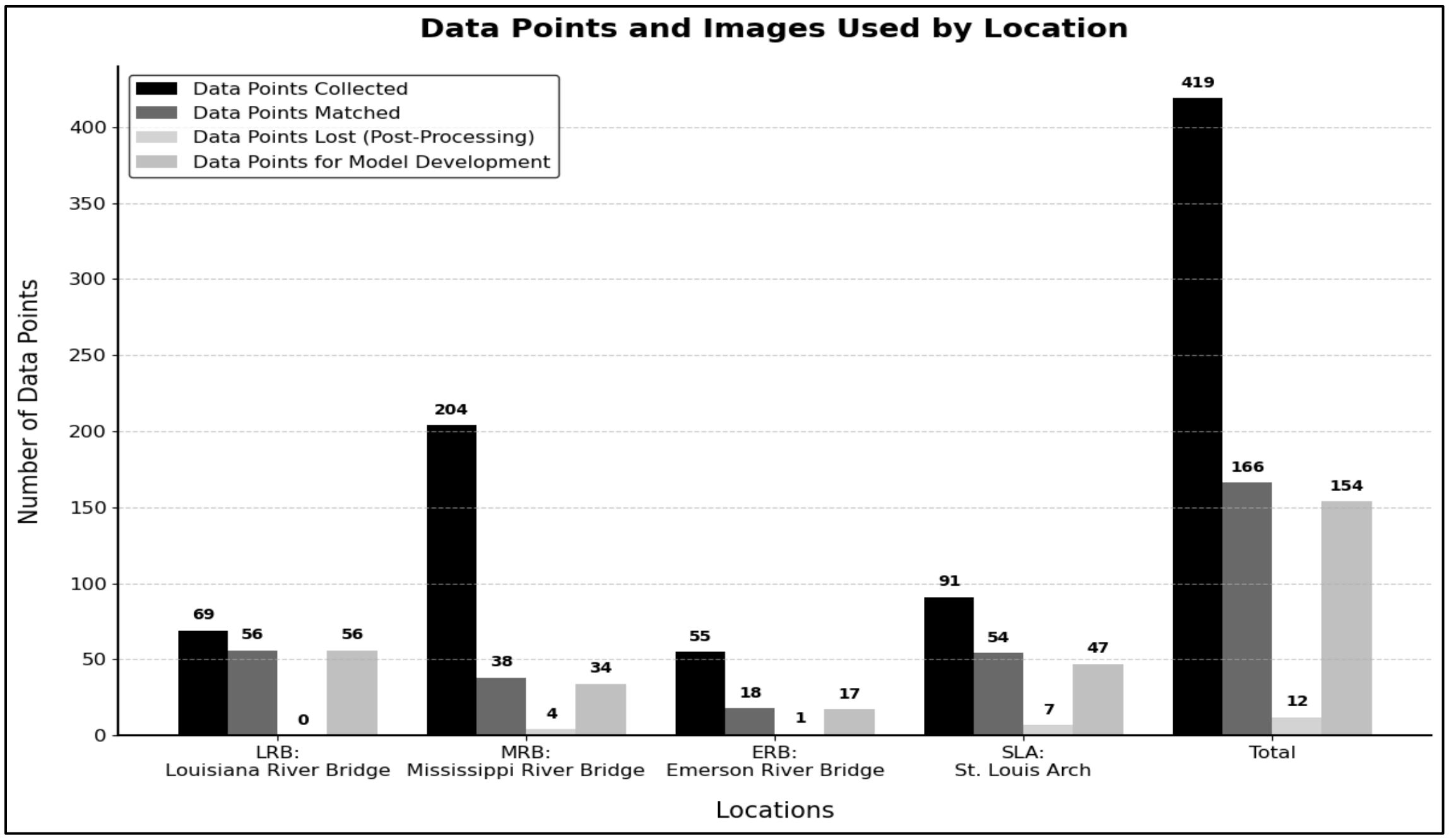

Figure 7: Distribution of Data Points and Images Used for Model Development by Location

### *4.3. Data Cleaning and Imputation*

Records with speeds over ground (SOG) less than 1 knot (indicating stationary vessels) (*A Method for Risk Analysis of Ship Collisions with Stationary Infrastructure Using AIS Data and a Ship Manoeuvring Simulator*, n.d.) or greater than 25 knots (indicating erroneous data) (*Resistance Characteristics of Semi-Displacement Mega Yacht Hull Forms*, n.d.) except for 102.3 knots (denotes vessel speed is unavailable) (*AIS Fundamentals*, n.d.) were removed. Records of statuses 1 (at anchor) and 2 (not under command) were also removed. Vessels at anchor are stationary and vessels not under command may have abnormal trajectories, both of which do not reflect typical vessel movement patterns. These steps reduced the dataset from 166 to 154 records.

Since several of the fields available in the AIS data are manually input, values can be missing (Liu & Ma, 2022). Therefore, vessel length, width, and draft were imputed. A K-means (Krishna & Narasimha Murty, 1999) clustering algorithm was used for imputation (Patil et al., 2010), grouping similar records based on SOG, vessel type, cargo type, status, time spent at various speeds, and standard deviations of SOG and acceleration. Seven clusters were determined optimal through iterative testing, balancing specificity and model complexity. Missing length, width, and draft values were imputed based on the characteristics of each cluster.

### *4.4. Data Augmentation*

Class imbalance in barge presence was addressed by upsampling the minority class using a modified Synthetic Minority Over-sampling Technique (SMOTE) and downsampling the majority class by removing records in excess of three (e.g., if there were 8 records of vessels with 10 barges, we kept only 3 randomly) **(Figure 8**). The minority class was weighted three times more than the majority class during training. 70% of the real data was used for training and validation, and 30% for testing. The synthetic data was included in the training set.

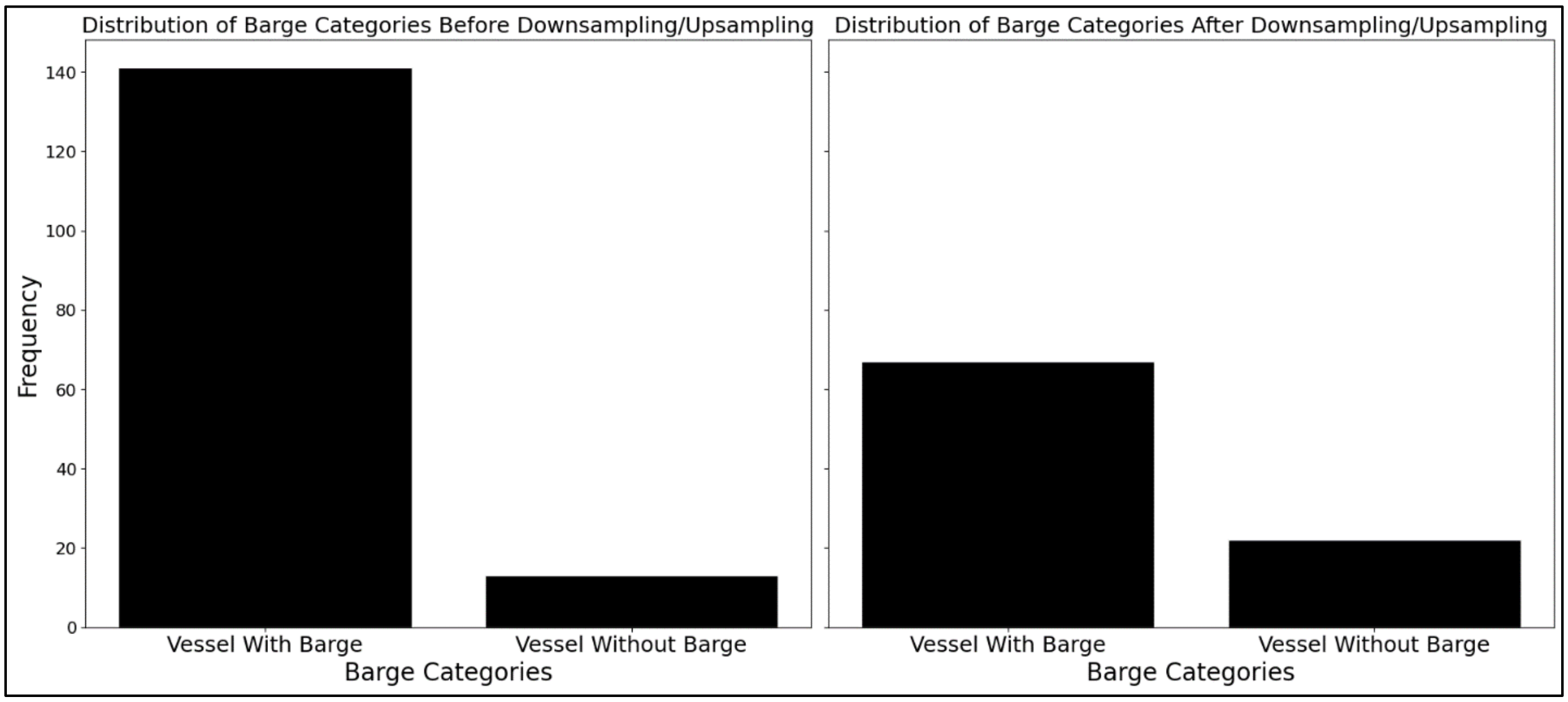

Figure 8: Effect of resampling on the distribution of barge categories

To address class imbalance (**Figure 9A**) in predicting barge quantities, particularly for minority classes (more than three barges), a modified SMOTE (Chawla et al., 2002) was developed. The algorithm identified the nearest neighbors for each minority class sample and interpolated between them to create new synthetic samples. For continuous features, random values within the sample-neighbor range were selected, while the mode of nearest neighbors was used for categorical features. Constraints, such as maintaining logical quartile order, ensured the validity of synthetic samples. The algorithm generated an overall increment of 29% by adding 50% more samples for imbalanced classes. The synthetic samples were combined with the original data to create an augmented dataset, resulting in a more balanced class representation and improved training for machine learning models **(Figure 9B**).

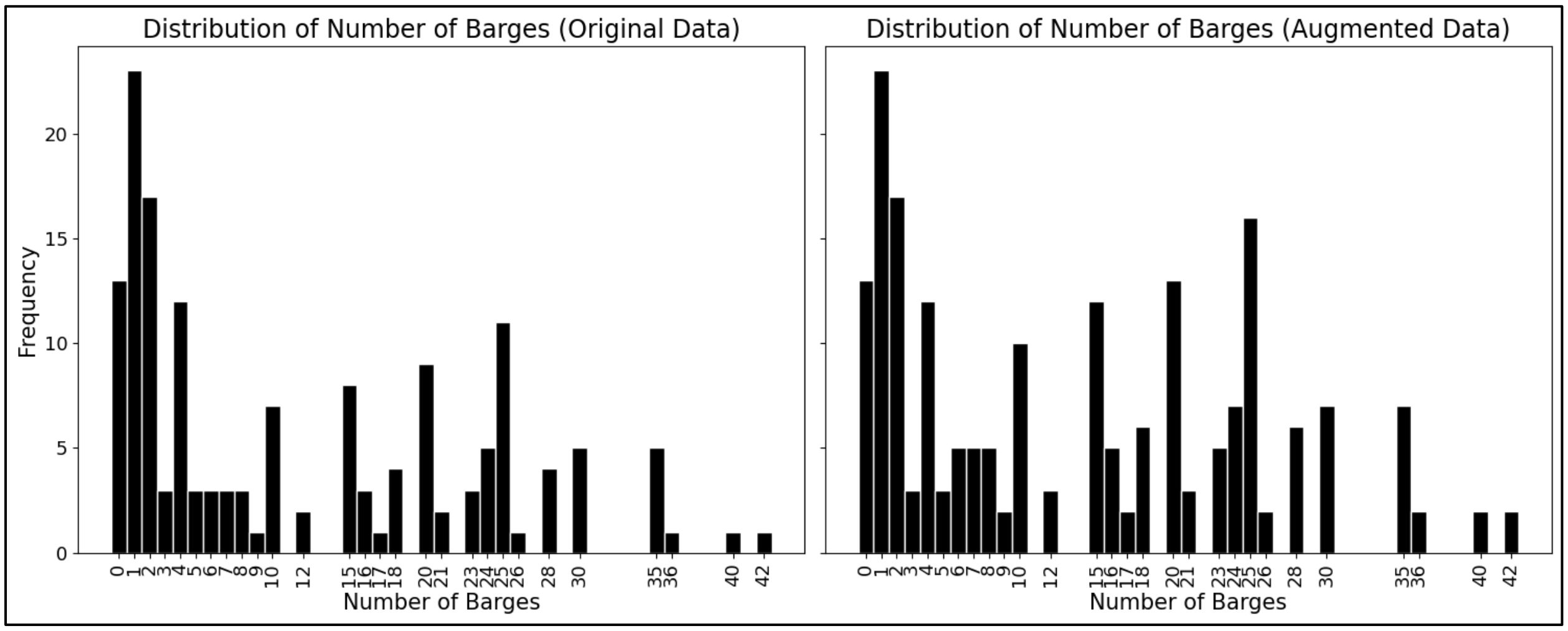

Figure 9: Distribution of Sample Before and After SMOTE Augmentation

For predicting the number of barges, the dataset was split into training and validation (85%) and test (15%) sets, with stratification based on the target variable to maintain class distribution. The training and validation set included 85% of real data and synthetic data generated by the custom SMOTE algorithm to balance minority class representation. The test set, consisting solely of real-world data, was kept separate throughout the model training process to accurately measure the model's predictive capabilities in real world scenarios. The distribution of classes for predicting barge quantities (including augmentation) is shown in **Figure 10.**

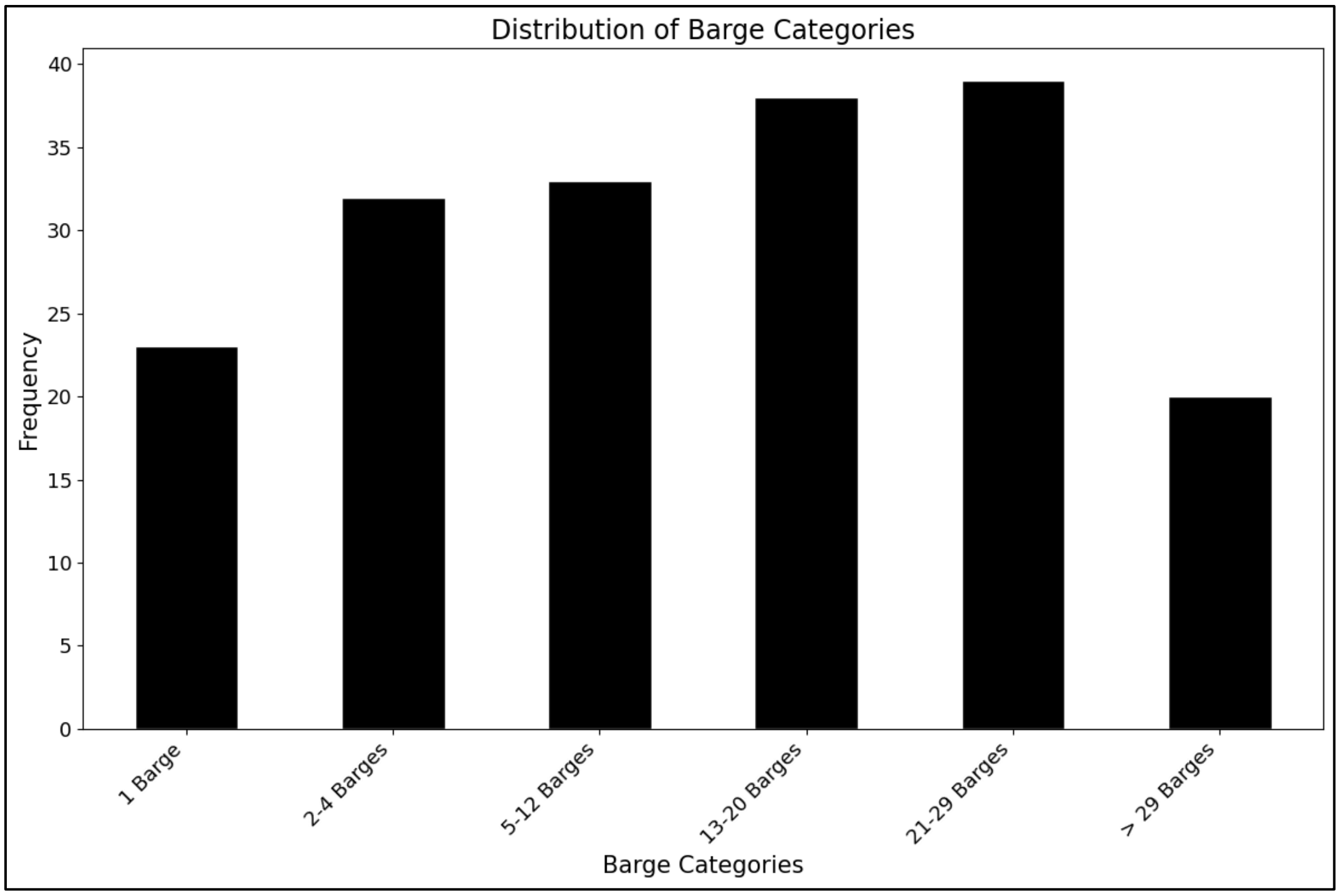

Figure 10: Distribution of different barge categories for predicting barge quantities

The hyperparameters tuned for each model, their tuning ranges, and the selected values for barge quantities are shown in **Table 2**.

Table 2: Hyperparameter Ranges and Selected Values for Machine Learning Models

| **Model** | **Hyperparameter** | **Description** | **Tuning Range** | **Selected Value** |
|---|---|---|---|---|
| **Barge Presence** | | | | |
| Random Forest | n_estimators | Number of trees in the forest | [50 - 300] | 200 |
| | max_depth | Maximum depth of the tree | [5 - 50] | 5 |
| | min_samples_split | Minimum number of samples required to split an internal node | [2 - 15] | 5 |
| | min_samples_leaf | Minimum number of samples required to be at a leaf node | [1 - 6] | 2 |
| LigtGBM | n_estimators | Number of boosting rounds | [50 - 300] | 50 |
| | learning_rate | Step size shrinkage | [0.01 – 1.0] | 1.0 |
| AdaBoost | min_child_samples | Minimum data points per leaf | [20 – 50] | 20 |
| | n_estimators | Number of boosting rounds | [50 - 300] | 50 |
| | learning_rate | Step size shrinkage | [0.01 – 1.0] | 1.0 |
| **Barge Quantity** | | | | |
| Random Forest | n_estimators | Number of trees in the forest | [50 - 300] | 100 |
| | max_depth | Maximum depth of the tree | [5 - 50] | 10 |
| | min_samples_split | Minimum number of samples required to split an internal node | [2 - 15] | 2 |
| | min_samples_leaf | Minimum number of samples required to be at a leaf node | [1 - 6] | 1 |
| LightGBM | n_estimators | Number of boosting rounds | [50 - 300] | 50 |
| | max_depth | Maximum depth of a tree | [(-1) - 7] | -1 |
| | learning_rate | Step size shrinkage | [0.01 – 1.0] | 0.1 |
| | num_leaves | Maximum number of leaves in one tree | [31 - 100] | 31 |
| | min_child_samples | Minimum data points per leaf | [20 – 50] | 20 |
| AdaBoost | n_estimators | Number of boosting rounds | [50 - 300] | 200 |
| | learning_rate | Step size shrinkage | [0.01 – 1.0] | 0.5 |

### *4.5. Model 1: Barge Presence*

Among the 10 models tested, five resulted in F1 scores above 93%, although they all used different feature sets. Of these five models, those that used the fewest number of features (less than 10) are described in this section. The fewer the number of features, the less complex the model, given that the performances remain the same. AdaBoost used 4 of the 36 features and achieved an F1 score of 0.932 (**Table 3**, **Figure 11**). The selected features included Turning Measures ('NROT'), Speed Measures ('Acc_SD'), and interaction terms ('Len*SOG_Q2', 'Wid*SOG_Q2').

| Table 3: Performance Metrics of Various Models for Predicting Barge Presence Model | **F1 Score** | **Recall** | **Precision** | **Accuracy** | **ROC-AUC** | **No. Features** | **Selected Features** |
|---|---|---|---|---|---|---|---|
| AdaBoost | 0.932 | 0.975 | 0.900 | 0.958 | 0.956 | 4 | 'NROT', 'Acc_SD', 'Len * SOG_Q2', 'Wid * SOG_Q2' |
| RF + AdaBoost | 0.932 | 0.975 | 0.900 | 0.958 | 0.975 | 7 | 'NROT', 'SOG_SD', 'Acc_SD', 'Wid * SOG_Q2', 'Acc_SD * SOG_SD', '(SOG_SD)^2', '(Acc_SD)^2' |
| RF + AdaBoost + Light GBM (Ensemble) | 0.932 | 0.975 | 0.900 | 0.958 | 0.975 | 7 | 'NROT', 'SOG_SD', 'Acc_SD', 'Wid * SOG_Q2', 'Acc_SD * SOG_SD', '(SOG_SD)^2', '(Acc_SD)^2' |

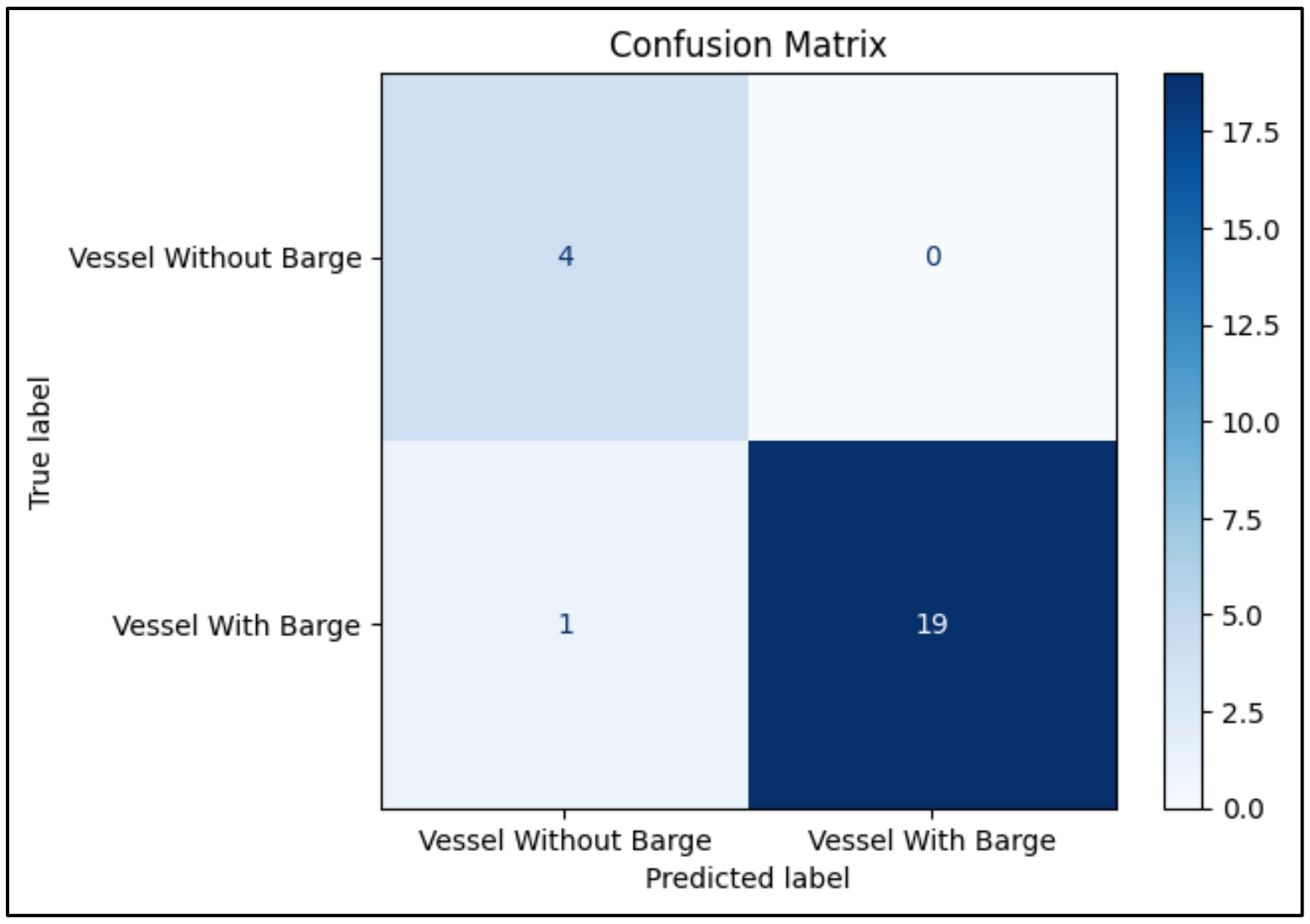

Figure 11: Confusion Matrix for the AdaBoost Model for Predicting Barge Presence

### *4.6. Model 2: Barge Quantity*

The ensemble model combining RF and AdaBoost performed best with an F1 score of 0.886 (**Table 4**), showing the lowest number of misclassifications **(Figure 12)** as compared to the nine other models. This model used 35 out of the 36 features **(Table 4).** The feature not included was 'Status_Other'. For simplicity, the top three models are shown.

Table 4: Performance Metrics of Machine Learning Models for Barge Count Prediction

| **Model** | **F1 Score** | **Recall** | **Precision** | **Accuracy** | **ROC - AUC** | **No. Features** | **Selected Features** |
|---|---|---|---|---|---|---|---|
| RF + AdaBoost (Ensemble) | 0.886 | 0.883 | 0.911 | 0.884 | 0.948 | 35 | 'PTST_SOG_5.4_to_6.9', 'Acc_SD', 'Len', 'Len*Wid', 'Len*SOG_Q2', 'Wid*SOG_Q2', 'Acc_SD*SOG_SD', 'Len*WidSOG_Q2', |

| **Model** | **F1 Score** | **Recall** | **Precision** | **Accuracy** | **ROC - AUC** | **No. Features** | **Selected Features** |
|---|---|---|---|---|---|---|---|
| | | | | | | | '(Len)^2', '(SOG_SD)^2', '(Acc_SD)^2', '(Len)^3', 'NROT', 'Len*Wid*SOG_Q2', 'SOG_Q1', 'SOG_Q3', 'VDraft', 'Wid', '(Wid)^2', 'Status_0', '(SOG_Q2)^2', 'SOG_Q2', '(SOG_Q2)^3', 'Status_12', 'CT_52', 'CT_32', 'VT_31_Towing', 'CT_31', 'VT_52_Tug', 'CT_57', 'CT_Other', 'VT_Other', 'Status_15' |
| RF | 0.862 | 0.883 | 0.869 | 0.864 | 0.946 | 12 | 'PTST_SOG_5.4_to_6.9', 'Acc_SD', 'Len', 'Len*Wid', 'Len*SOG_Q2', 'Wid*SOG_Q2', 'Acc_SD*SOG_SD', 'Len*WidSOG_Q2', '(Len)^2', '(SOG_SD)^2', '(Acc_SD)^2', '(Len)^3' |

| **Model** | **F1 Score** | **Recall** | **Precision** | **Accuracy** | **ROC - AUC** | **No. Features** | **Selected Features** |
|---|---|---|---|---|---|---|---|
| RF + Light GBM (Ensemble) | 0.836 | 0.842 | 0.858 | 0.818 | 0.951 | 18 | ‘PTST_SOG_5.4_to_6.9’, ‘Acc_SD’, ‘Len’, ‘Len*Wid’, ‘Len*SOG_Q2’, ‘Wid*SOG_Q2’, ‘Acc_SD*SOG_SD’, ‘Len*WidSOG_Q2’, ‘(Len)^2’, ‘(SOG_SD)^2’, ‘(Acc_SD)^2’, ‘(Len)^3’, ‘SOQ_Q1’, ‘SOG_Q3’, ‘NROT’, ‘PTST_SOG_<5.4’, ‘SOG_SD’, ‘VDraft’, ‘Acc_SDSOG_SD’ |

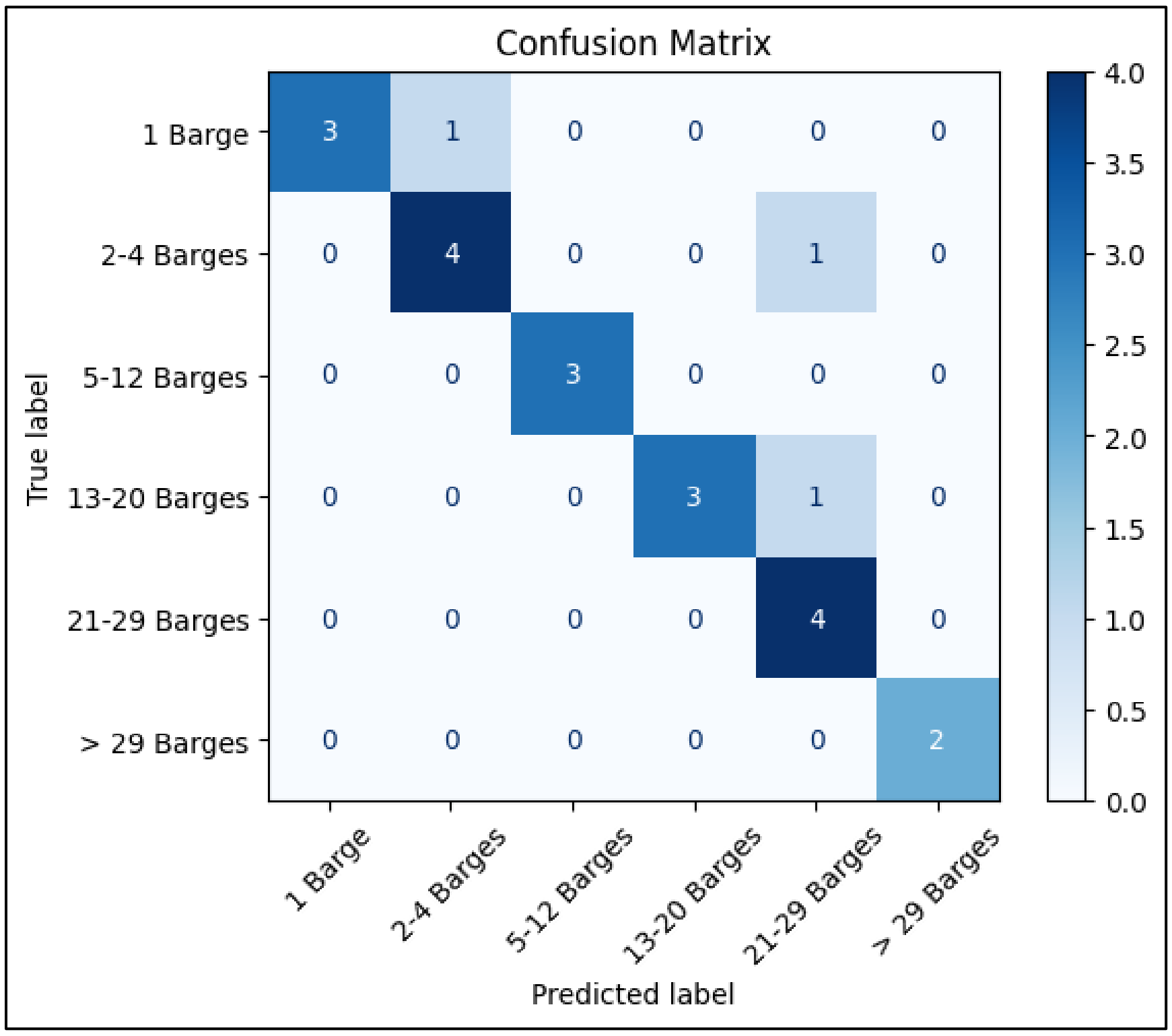

Figure 12: Confusion Matrix for RF + AdaBoost Ensemble Model

### *4.7. Sensitivity Analysis*

#### *4.7.1. Optimal River Segment Length*

A sensitivity analysis was conducted to determine the optimal river segment length for recreating vessel trajectories. Various segment sizes, ranging from 0.1 to 2 miles, were evaluated over a 382-mile stretch. Smaller segments provided higher accuracy but increased computational costs. A 0.3-mile segment size struck a balance, resulting in an error of 1.26 miles over 382 miles **(Figure 13)**. This size effectively predicted vessel trajectories and arrival times at specific river locations, such as camera positions, without oversimplifying complex navigational behaviors. The fixed segment size was optimal for capturing intricate river segments, such as long curves, ensuring accurate trajectory representation across all river sections.

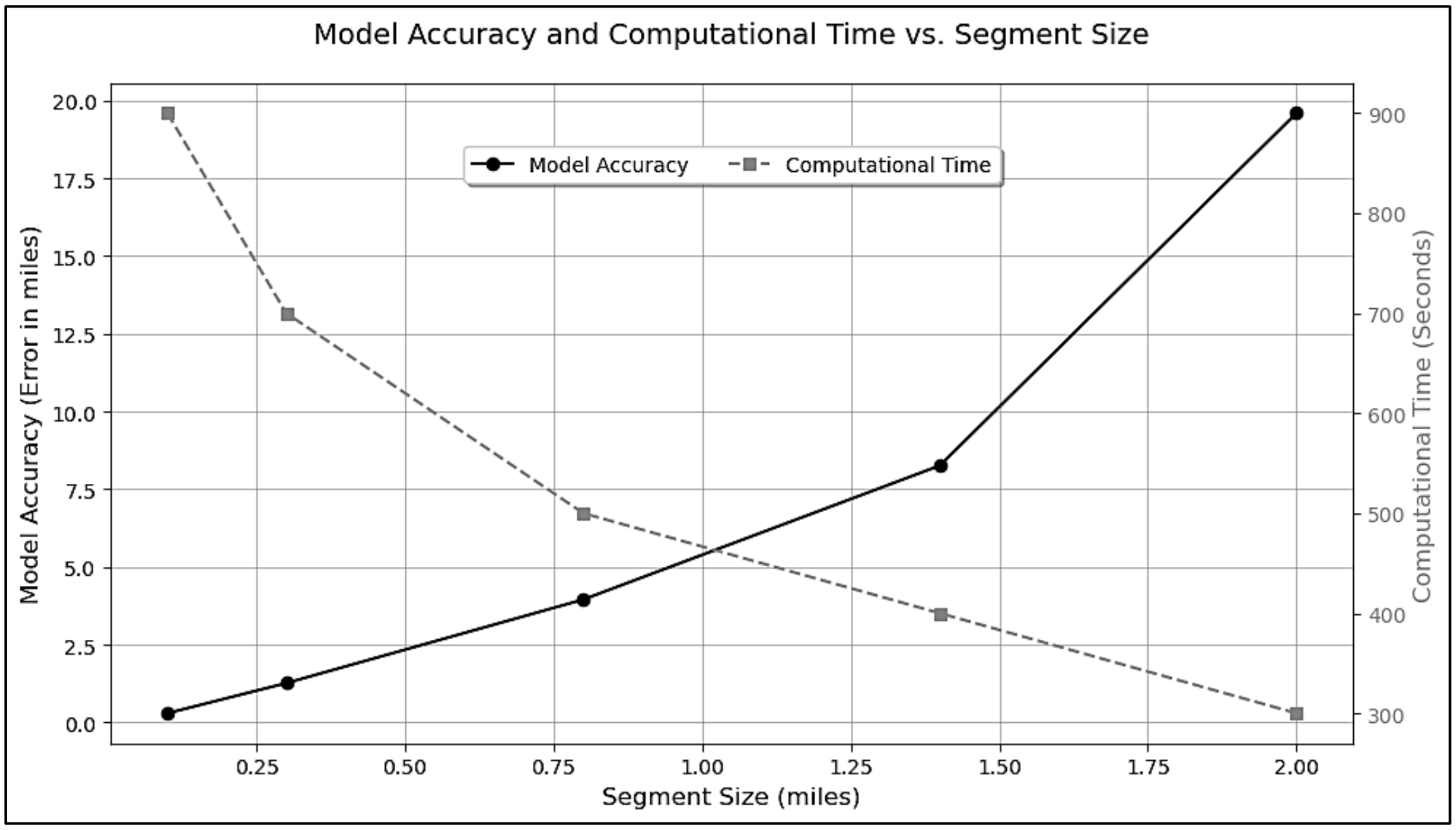

Figure 13: Sensitivity Analysis Results for Segment Size in Trajectory Prediction

#### 4.7.2. *Optimization of Barge Class Groupings*

In this paper, barge counts are represented as categorical values. A sensitivity analysis was conducted to find an optimal grouping of barge counts that enhanced predictive accuracy. First, models were initially trained to predict precise barge counts as continuous values (27 'classes'). A confusion matrix was generated to understand the distribution of misclassifications among the barge counts. To improve model training and maintain prediction granularity, classes with the lowest performance (high misclassification rates) were identified. Adjacent low-performing classes were combined. For example, counts of 1 and 2 were grouped together if frequently misclassified. The model's performance was iteratively evaluated, and groupings were refined until an optimal balance between performance and granularity was achieved. The strategy reduced the initial 27 classes to 6 categories **(Figure 14).**

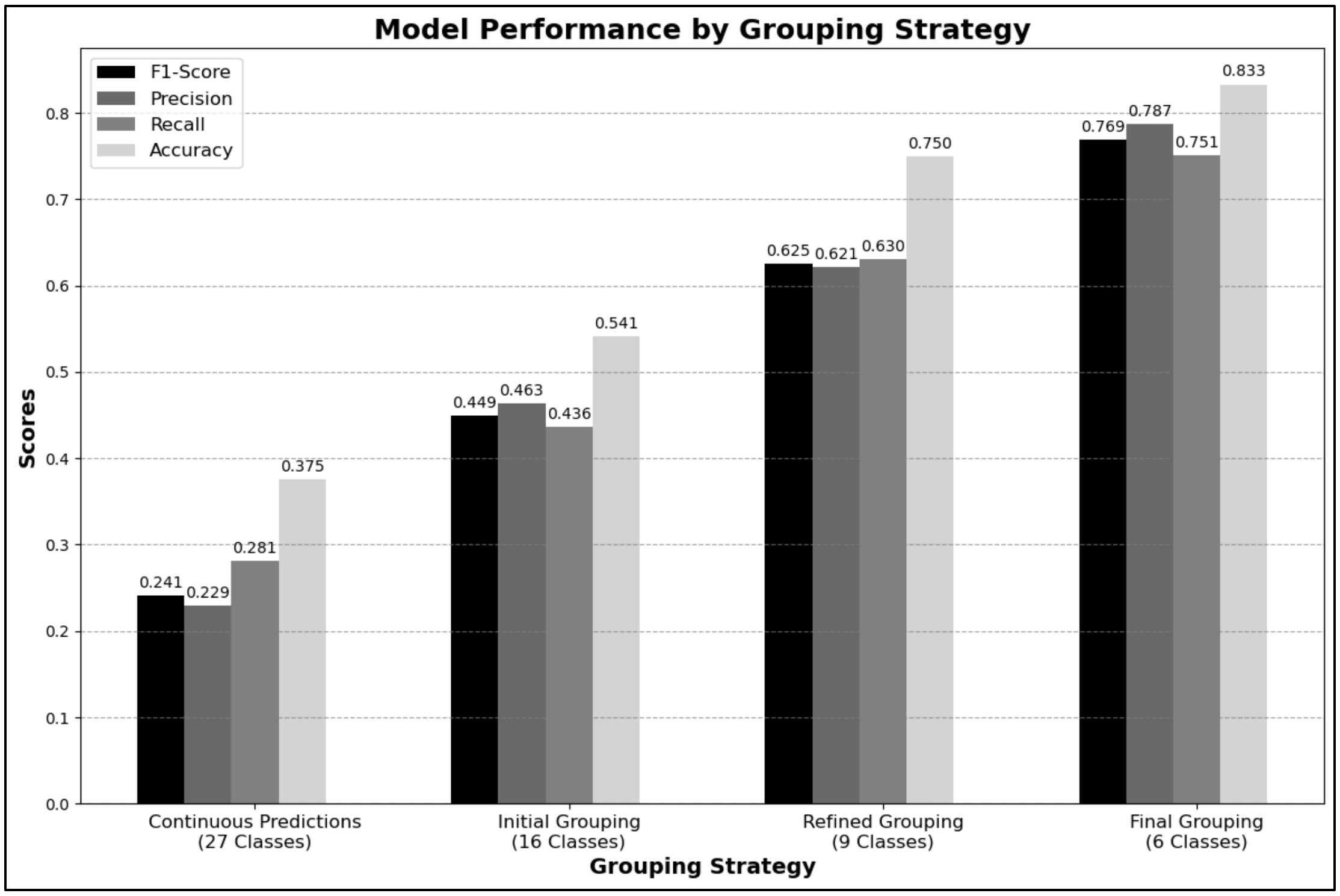

Figure 14: Model Performance by Grouping Strategy

***Spatial Transferability Analysis***

To evaluate the spatial transferability of the best model (RF + AdaBoost Ensemble) in predicting barge quantity, the model was trained on location-specific datasets from LRB, MRB, and SLA with augmentation and tested on ERB. The model achieved a test F1 score of 0.702, precision of 0.694, recall of 0.717, and accuracy of 0.688 using 12 features including 'Len*Wid', 'Len*SOG_Q2', 'Wid*SOG_Q2', 'Acc_SD*SOG_SD', 'Len*Wid*SOG_Q2', '(SOG_Q2)^2', '(Len)^2', '(Wid)^2', '(SOG_SD)^2', '(Acc_SD)^2', '(SOG_Q2)^3', '(Len)^3' **(Figure 15)**.

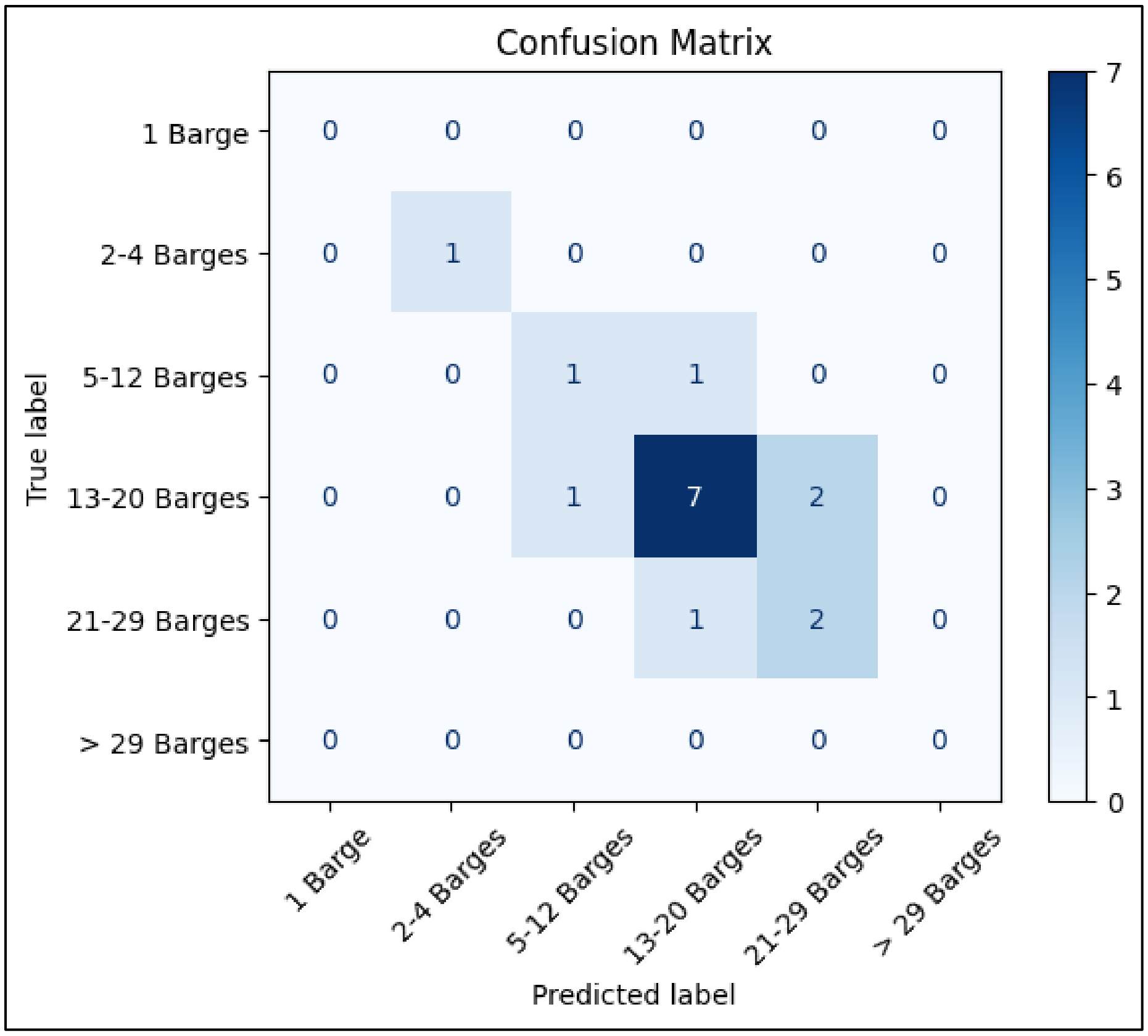

Figure 15: Confusion Matrix for Spatial Transferability Analysis

## 5. Conclusion

This study proposed a novel approach to predicting the presence and quantity of barges transported by vessels on inland waterways using AIS data and machine learning. 36 features derived from AIS data represented vessel maneuvering characteristics such as speed, rate of turn, and acceleration/deceleration, in addition to vessel characteristics such as width, length, and draft. Labeled sample data was uniquely created by matching labeled images taken from traffic cameras located on bridges crossing inland waterways to AIS records. This process required trajectory reconstruction of AIS records to establish timestamps at bridge/camera locations. Four camera/bridge locations were chosen to represent the Mississippi River waterway traffic generating 419 images in 15 months. Of these, 166 were matched to AIS records and used for model development and testing.

The prediction models include barge presence detection and barge count estimation. RF, SVM, KNN, Decision Trees, and NN as well as ensemble approaches including XGBoost, AdaBoost, and LightGBM were evaluated.

Bayesian Optimization was used for hyperparameter tuning. For detecting barge presence, the AdaBoost model achieved an F1 Score of 0.932 following a binary classification scheme: [Vessel with Barge; Vessel Without Barge]. Using RFE, 4 of the 36 features derived from AIS data were used. For barge count estimation, the RF + AdaBoost ensemble model achieved the highest accuracy with an F1 score of 0.886 following a six-class scheme: [1 Barge, 2-4, 5-12, 13-20, 21-29, and >29 (which ranges from 30-42 barges)].

One limitation is the reliance on manually collected data from traffic cameras, which is time-consuming and limits the number of samples available for model development. Furthermore, the camera locations may not fully represent the diversity of vessel movements across the entire inland waterway system. Expanding the study to include more camera locations could provide a more comprehensive understanding of barge movements. To mitigate the limitations associated with manual data collection, there is a need to transition towards automated methods. Manual methods were used because there were no datasets available for training a computer vision model to do automated counts. To this end, manual methods are crucial for creating training and validation datasets for developing robust computer vision models which are for predicting barge counts.

Lastly, the study primarily focused on the quantities of barges carried by vessels, without considering the types and quantities of commodities transported. Future work should aim to incorporate additional data on cargo types to provide more detailed insights into the flow of goods on inland waterways. This could be achieved by integrating data from other sources, such as LPMS and WCSC.

## Author contributions

The authors confirm their contributions to the paper as follows: study conception and design: Geoffery Agorku, Sarah Hernandez; data collection: Geoffery Agorku, Maria Falquez; analysis and interpretation of results: Geoffery Agorku, Shihao Pang, Sarah Hernandez, Subhadipto Poddar; draft manuscript preparation: Geoffery Agorku, Subhadipto Poddar.